# Deep Nearest Class Mean Model for Incremental Odor Classification

Yu Cheng[1,2], Kin-Yeung Wong[2], Kevin Hung[2], Weitong Li[1], Zhizhong Li[1], Jun Zhang[1]

[1]School of Information Engineering, Guangdong University of Technology

[2] School of Science and Technology, The Open University of Hong Kong

**Abstract:** In recent years, more machine learning algorithms have been applied to odor classification. These odor classification algorithms usually assume that the training datasets are static. However, for some odor recognition tasks, new odor classes continually emerge. That is, the odor datasets are dynamically growing while both training samples and number of classes are increasing over time. Motivated by this concern, this paper proposes a Deep Nearest Class Mean (DNCM) model based on the deep learning framework and nearest class mean method. The proposed model not only leverages deep neural network to extract deep features, but is also able to dynamically integrate new classes over time. In our experiments, the DNCM model was initially trained with 10 classes, then 25 new classes are integrated. Experiment results demonstrate that the proposed model is very efficient for incremental odor classification, especially for new classes with only a small number of training examples.

## I. Introduction:

Odor recognition with electronic nose (E-nose) plays an important role in many applications, such as detection and diagnosis in medicine [1], [2], [3], searching for drugs and explosives [4], [5], quality control in food producing chain [6], [7], and locating gas source [8]. Some odor-based classification algorithms (like Chinese herb recognition, medicine diagnoses and dangerous goods searching) are required to update their knowledge over time as they are introduced to odors of new classes. These tasks are considered as *incremental odor classification*, where the number of both samples and classes in training dataset gradually increases. To deal with incremental odor classification, non-incremental algorithms have to be retrained from scratch, while incremental learning algorithms only need to be updated when new classes are introduced. The difference between non-incremental and incremental learning is illustrated in Fig.1.

To deal with new classes in classification, one could consider model-free classifiers such as K-Nearest Neighbor (KNN) and Nearest Class Mean (NCM) classifiers. KNN is a highly nonlinear classifier that has the ability to achieve an accuracy rate comparable to much more complex methods [9]. It integrates new classes by simply adding new training samples (of new classes) to the database which can be used for classification directly. The cost of this simple processing is that it not only needs to store all training samples, but also has large computational burden for prediction. This restricts the application of KNN in many real-world situations. Compared to the KNN method, NCM is a much more efficient classifier [10]. However, NCM is just a linear classifier. It cannot be applied directly in some complex classification tasks (e.g. odor classification and vision-based pattern recognition) where the boundary of classes is nonlinear in data space.

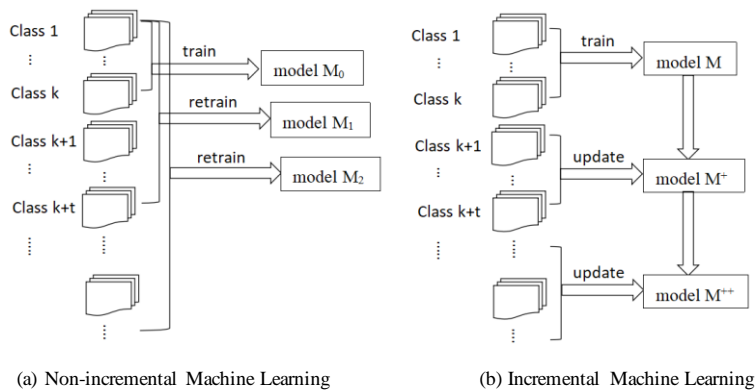

(a) Non-incremental Machine Learning  (b) Incremental Machine Learning

Fig.1 Different learning processes between non-incremental machine learning and incremental learning.

To improve the capability of NCM, T. Mensink et al. proposed a metric learning algorithm, named Nearest Class Mean Metric Learning (NCMML), where samples are enforced to be closer to their class mean than other class means in the projected space [11]. The metric learning projects high-dimensional images to low-dimensional space by linear projection. It is effective for high dimension samples (e.g. image), but is not fit for odor data, since the dimension of odor is low (about 5 to 10). M. Ristin et al. presented a new type of random forests based on NCM, called Nearest Class Mean Forests (NCMF), where the decisions at each node are formed by a small random subset of class means observed at that node. They also introduced strategies for updating forest structure in order to integrate new classes [12], [13]. NCMF needs large training images of new classes to train each node of the forest. Nevertheless, the amount of new classes' samples is relatively small in odor recognition because measuring an odor is much more difficult than taking a picture.

Besides the small amount of new samples, another difficulty of odor classification is that gas sensors are sensitive to physical environment (e.g. temperature, humidity and pressure), gas concentration and abnormal odors [14]. The data measured by gas sensors may include signal noise, shift, drift, etc. [14], [15]. These inaccurate attributes of gas sensor seriously hinder the application of E-nose in the actual environment. To solve this problem, deep learning algorithms may be an appropriate choice. Deep learning can extract deep features, improve discriminative aspects, and suppress the irrelevant variations of raw data [16]. Therefore, it helps reduce or eliminate signal noise, shift and drift of gas sensors, and could ultimately improve accuracy of odor recognition. This advantage of deep learning has been proven in many other areas such as speech recognition [17], [18], visual object recognition [19], [20], object detection [21] and natural language understanding [22], [23].

If deep neural network is used for incremental odor classification, two problems need to be solved. First, the existing deep neural networks cannot efficiently handle the dynamically growing dataset where not only the amount of training samples, but also the number of classes increases over time. This means they have to be trained from scratch when the training dataset are added with new classes. Second, deep neural network generally requires large training dataset to perform well. However, in incremental odor recognition, the dataset of new classes are usually small. In order to deal with these two problems, we propose the deep nearest class mean (DNCM) model which modifies not only the structure but also the training algorithm of deep neural network. DNCM embeds a nearest class mean model as a layer (NCM layer) in deep neural network (see Fig. 2). The NCM layer can effortlessly incorporate new classes by adding means of new class, so it can tackle the first difficulty. To handling the second problem, we divide the whole training procedure into two phases. In the first phase, to ensure the hidden layers of DNCM can extract the right features, a relatively large training dataset is used to train the DNCM. Then, the parameters of the hidden layers are fixed from this point onwards and are shared in the sequent phase. In the second phase, the samples of new class are only used to train the parameters of NCM layer, and are not used to train the hidden layers. The parameters of NCM layer are small, since they are only the means of each class. So, although dataset of new classes are relatively small, the model's generalization accuracy will not significantly deteriorate. The details of the model will be discussed in Section III, and the details of training strategies will be discussed in Section IV.

The contributions of this paper can be summarized as follows. First, we extend a deep learning model to an incremental learning model, Deep Nearest Class Mean (DNCM), which can continually integrate new classes over time. It does not need to be retrained from scratch. Second, we introduce a loss function of DNCM to minimize the intra-class compactness and simultaneously maximize the inter-class distance in feature space. Third, we propose a two-phase learning method to train the DNCM model.

The rest of this paper is organized as follows. Section II reviews some related works. Section III describes the detail of the proposed model, and the training strategies are discussed in Section IV. Section IV presents the experimental results, analysis, and comparisons with the state-of-art methods in incremental odor recognition. Finally, the conclusion is given in Section VI.

## II. Related Works

In this section we review related works on odor recognition, incremental learning and transfer learning. With the development of artificial intelligence, more and more machine learning algorithms were proposed for odor recognition. L. Zhang et al. proposed discriminative subspace learning which targeted odor recognition across multiple E-nose [24], an extreme learning machine based self-expression for abnormal odor detection [25], and efficient algorithms for solving three key issues of E-nose technology: Discreteness, Drift, and Disturbance [14] and Domain Adaptation Extreme Learning Machines to deal with gas sensor drift [15]. Rodriguez et al. used support vector machine (SVM) for the calibration of gas sensor arrays [26]. Dixon et al. [27] compared the performance of five pattern recognition approaches in odor recognition: Euclidean Distance to Centroids (EDC), Linear Discriminant Analysis (LDA), Quadratic Discriminant Analysis (QDA), Learning Vector Quantization (LVQ) and Support Vector Machines (SVM). However, most previous odor classification methods assumed that the training dataset was static.

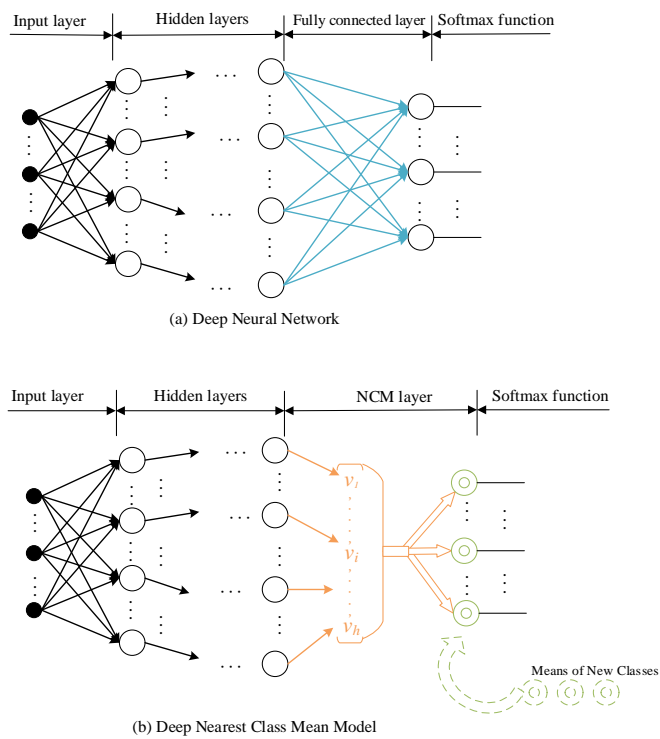

Fig. 2 (a) is deep neural network model which is generally composed of four parts: input layer, hidden layers, fully connected layer and a softmax function. (b) is the proposed deep nearest class mean (DNCM) model which replaces the fully connected layer by a NCM layer. The NCM layer treats the output of the last hidden layer as a feature vector $[v_1, \cdots, v_h]^T$. The solid olive concentric circles represent the mean (center) of classes that have already been learned by the model. The dotted olive concentric circles represent the mean (center) of new classes. The output of NCM layer is the distance vector between a feature vector and mean of classes.

To handle dynamically growing datasets, model-free method is one of competent choices. The classic model-free methods include K-Nearest Neighbor (KNN) [9], Kernel Density Estimators (KDE) [28], and Nearest Class Mean classifier (NCM) [10]. KNN conducts classification tasks by first calculating the distance between the test samples and all the training samples to obtain its nearest neighbors and then conducting its class. KNN has the difficulty to deal with large dataset. One reason is that it needs large storage space to save all the samples. Another problem is that although no effort is required for training, conducting the classification needs a large computational burden for each test sample. The Nearest Class Mean (NCM) classifier can be seen as a variant of KNN. Instead of saving all samples, it represents classes by the mean (center) of their samples. Compared to KNN, NCM needs less storage space and computation. However, it is a linear classifier, which is not fit for highly non-linear classification tasks (e.g. odor classification).

Incremental learning, as defined in [11], [12], [13], [29], [30], is the scenario where training data is not provided uniformly at random, but where classes are provided in sequence. As far as we know, the first work to deal with incremental odor recognition was proposed by Tudu et al. [7], where they used a set of fuzzy rules to evaluate black tea quality. Their method assume that there are small differences between the odor samples in the same class. Black tea satisfies this assumption, but many other types of odor, such as herbs, fruits and breath, do not. It means that rule-based methods may just be applied to a relatively small area.

Using multiple simple classifiers is another framework in incremental learning. One vs. rest [31], one vs. one [32] and Random forest [33] belong to this category. The strategy of one vs. rest involves training a single binary classifier (e.g. SVM) per class, with samples of the class as positive samples and all other samples as negatives. Suppose a model has already accommodated $N$ classes, when adding a new class $N+1$, it trains one new one vs. all classifier and updates their $N$ previous trained classifiers. The computational cost of training a new classifier and updating the $N$ trained classifiers is high [34]. The one vs. one framework trains $N(N-1)/2$ binary classifier for the $N$-way multi-class classification problem. To extend the model to the $(N+1)$-th class, it should train $N$ new binary classifiers. When the number of class is large, for example 25 classes, it needs to maintain 600 sub-classifiers. Therefore, the one vs. one framework is only suitable for scenarios which involve very small number of classes. Random forest uses hierarchical binary classifier. In [12], Marko Ristin et al. combined NCM and the incremental learning mechanism into random forest models. They proposed four methods for incrementally learning an NCM Forest: i) updating leaf statistics, ii) incrementally growing tree, iii) re-training subtree, and iv) reusing subtree. While the first three methods work with general type of node classifier, the last one uses the advantage of NCM for incremental learning. In addition to the above four methods, in [13], Marko Ristin et al. further proposed node sampling for partial tree update which prefers to update subtrees with poor performance. As an ensembles framework, random forest is a simple and flexible predictive model. Higher-accurate random forest requires more trees, more node binary classifiers, and more data. When the amount of training samples is not adequate, the accuracy of the random forest will drop significantly. As such, the NCMF is mainly used in the context of large scale training dataset.

Incremental learning is also related to transfer learning, which extracts knowledge from one or more source tasks and applies the knowledge to a target task [35], [36], [37], [38]. In the E-nose community, transfer learning mainly focuses on handling discreteness, drift, and disturbance of gas sensor [14], [15]. Deep neural network is also a commonly used tool for transfer learning. W. Ge et al. used a subset of training images from the original source to learn task whose low-level characteristics were similar to those from the target learning task, and jointly fine-tune shared convolutional layers from both tasks [39]. Y. Wu and Q. Ji [40] introduced a constrained deep transfer feature learning method to perform simultaneous transfer leaning and feature learning by performing transform learning in a progressively improving feature space iteratively. Some incremental learning methods were built on transfer learning. The MULticlass Transfer Incremental LEarning (MULTIpLE) [30] is an incremental learning method based on transfer learning and the multiclass extension of Least-Squares Support Vector Machine (LSSVM) [41]. It casts transfer learning as a constraint for the classifier of the *N+1* target class to be close to a subset of the *N* source classifiers.

Our work is mainly inspired by transfer learning which uses deep learning method such as deep feature extractor [39], [40]. On the one hand, the proposed DNCM leverages deep neural network to extract features from gas sensor signals in order to deal with the instability of gas sensors. On the other hand, it can seamlessly integrate new odor classes rather than retraining from scratch. The proposed DNCM takes advantage of the strengths of deep neural network and nearest class mean method, while eliminating their limitations.

### III. The Proposed Deep Nearest Class Mean (DNCM) Model

In this section, before introducing the proposed Deep Nearest Class Mean (DNCM) model, we briefly describe the concept of deep neural network and NCM classifier, and how to apply them in odor classification.

#### A. Deep Neural Network (DNN)

Deep Neural Network (DNN) uses an expressive hierarchical structure, i.e. multiple layers. Each layer contains multiple neurons which connect with the neurons of its neighbor layers' neurons. The output $\boldsymbol{\alpha}_i$ of the *i-th* layer is the nonlinear transform of the linear composition of the previous layer's output $\boldsymbol{\alpha}_{i-1}$.

$$\boldsymbol{\alpha}_i = \phi_i(W_{i-1} \times \boldsymbol{\alpha}_{i-1}) \tag{1}$$

where $W_{i-1}$ is the weight matrix connects *(i-1)-th* layer and *i-th* layer.

In odor classification task, the output of DNN model is probability of input odor belong to each class, as presented in equation (2):

$$g(\boldsymbol{\alpha}_i) = \frac{[e^{\alpha_{h1}}, e^{\alpha_{h2}}, \cdots, e^{\alpha_{hK}}]^T}{\sum_{k=1}^{K} e^{\alpha_{hk}}} \tag{2}$$

Where $\boldsymbol{\alpha}_h = [\boldsymbol{\alpha}_{h1}, \boldsymbol{\alpha}_{h2}, \cdots, \boldsymbol{\alpha}_{hK}]^T$ is the output of the full connected layer.

#### B. Nearest Class Mean (NCM) classifier

The key idea of NCM is that each class is represented by its mean (center) of its elements [10].

$$\boldsymbol{c}_k = \frac{1}{N_k} \sum_{\boldsymbol{v}_i \in class\ k} \boldsymbol{v}_i \tag{3}$$

$$k^* = \underset{k^* \in \{1,\cdots,K\}}{\operatorname{argmin}} d(\boldsymbol{v}_i, \boldsymbol{c}_k) \tag{4}$$

where $\boldsymbol{c}_k$ is the mean of each class $k$, and $\boldsymbol{v}_i$ is a feature vector, and $N_k$ is the number of samples in class $k$, and $d(\boldsymbol{v}_i, \boldsymbol{c}_k)$ is the distance metric between $\boldsymbol{v}_i$ and class mean $\boldsymbol{c}_k$. NCM assigns $\boldsymbol{v}_i$ to the class $k^*$ with the closest mean.

*C. Deep Nearest Class Mean Model*

In this paper, we consider the incremental multi-class odor recognition scenario where new classes gradually become available. The proposed Deep Nearest Class Mean (DNCM) provides a framework to extend a deep neural network (DNN) model to an incremental learning model. DNN can be decomposed into four parts: input layer, hidden layers, fully connected layer, softmax function (see Fig. 2(a)). In contrast to DNN that adopts a fully connected layer, the proposed DNCM use a NCM layer instead (see Fig. 2 (b)). In DNCM, the hidden layers are responsible for transforming raw gas data into feature vectors. When new classes are available, the NCM layer integrates them by adding the mean of new class, rather than retraining. The softmax functions are used to compute the probability of each class to which the odor may belong.

More specifically, we perform the following procedure to predict a new odor, $\boldsymbol{x}_i$, with the DNCM. Hidden layers are used as deep feature extractor $f(\cdot)$ to compute the feature vector $\boldsymbol{v}_i$ of $\boldsymbol{x}_i$:

$$\boldsymbol{v}_i = f(\boldsymbol{x}_i) = \phi_{h-1}(W_{h-2} \times \phi_{h-2}(\cdots \phi_1(W_0 \times \boldsymbol{x}_i))\cdots) \tag{5}$$

, where $\phi_i(\cdot)$ is a non-linear function of the *i-th* hidden layer. And $W_{i-1}$ is the weight matrix connects *(i-1)-th* hidden layer and *i-th* hidden layer.

Then NCM layer will output the distance, $d_{ik}$, between $\boldsymbol{v}_i$ and the center of each class $\boldsymbol{c}_k$

$$d_{ik} = d(\boldsymbol{v}_i, \boldsymbol{c}_k) \tag{6}$$

, where $d(\boldsymbol{v}_i, \boldsymbol{c}_k)$ is the distance function between $\boldsymbol{v}_i$ and $\boldsymbol{c}_k$. The distance $d_{ik}$ will be translated into a probability form by softmax function, denoted as $p(\boldsymbol{c}_k|\boldsymbol{v}_i)$:

$$p(\boldsymbol{c}_k|\boldsymbol{v}_i) = \frac{e^{-d_{ik}}}{\sum_{l=1}^{K} e^{-d_{il}}} \tag{7}$$

DNCM will assign the odor $\boldsymbol{x}$ to the class $k^* \in \{1, \cdots, K\}$ with the largest probability.

$$k^* = \underset{k \in \{1,\cdots,K\}}{\operatorname{argmax}} p(\boldsymbol{c}_k|\boldsymbol{v}_i) \tag{8}$$

Further, we introduce the loss function of DNCM:

$$\mathcal{L}(\mathbb{D}) = -\sum_{i=1}^{n} \sum_{k=1}^{K} t_{ik} \log p(\boldsymbol{c}_k|\boldsymbol{v}_i) \tag{9}$$

, where $\mathbb{D} = \{(\boldsymbol{x}_1, y_1), \cdots, (\boldsymbol{x}_n, y_n)\}$ is the training dataset, and $t_{ik}$ is the *k-th* element of vector $\boldsymbol{t}_i$ which is a one-hot vector corresponding to $y_i$. The optimal parameters $\mathbf{W} = [W_0, \cdots, W_h]$ of DNCM is the solution of $\underset{\mathbf{W}}{\operatorname{argmin}} \mathcal{L}(\mathbb{D})$. Substituting equation (6), (7) into equation (9) yields the following equation:

$$\underset{\mathbf{W}}{\operatorname{argmin}} \mathcal{L}(\mathbb{D}) = -\sum_{i=1}^{n}\sum_{k=1}^{K} t_{ik} \log \frac{e^{-d(v_i, c_k)}}{\sum_m e^{-d(v_i, c_m)}} \quad (10)$$

The object of equation (10) is searching parameter matrix $\mathbf{W}$ to minimize $d(\boldsymbol{v}_i, \boldsymbol{c}_{y_i})$ (the distance between $\boldsymbol{v}_i$ and mean of its own class) and simultaneously maximize $\sum_k d(\boldsymbol{v}_i, \boldsymbol{c}_k)$ (the sum of distance between $\boldsymbol{v}_i$ and each mean of classes). So, optimizing loss function is equivalent to minimizing intra-class compactness and maximizing the inter-class separability simultaneously in feature space.

**IV Incremental Learning Strategies for DNCM**

In this section, we will describe the training method for DNCM. The learning procedure can be divided into two phases: initial training and updating training. In initial training phase, a relatively large dataset is used to train the DNCM. The parameters of DNCM' hidden layers are learned in this phase. In updating training phase, the DNCM constantly integrate new classes to handle dynamically growing dataset.

*A. Initial training*

In initial training phase, there is a relatively large initial training dataset $\mathbb{D}^\Delta = \{(\boldsymbol{x}_1^\Delta, y_1^\Delta), (\boldsymbol{x}_2^\Delta, y_2^\Delta), \dots, (\boldsymbol{x}_n^\Delta, y_n^\Delta)\}$ composed of $n$ labeled samples $(\boldsymbol{x}_i^\Delta, y_i^\Delta)$, where $\boldsymbol{x}_i^\Delta$ is a signal of gas sensor of an odor at one moment, and $y_i^\Delta \in \{1, 2, \dots, K\}$ is a class label which denotes the class that $\boldsymbol{x}_i^\Delta$ belongs to. The initial dataset is static which mean that it is available before training. The parameters of hidden layers will be learned in initial training phase.

Back Propagation (BP) algorithm is used to train the DNCM in initial training phase. However, because the loss function of DNCM contains the mean of classes (see equation (10)), which will be changed along with the update of the parameters of DNCM at each epoch, therefore, the main difference between the procedure adopted in the proposed method and the one used in the deep neural networks is that the mean of classes is recomputed at each epoch. The initial training procedure performs the following steps iteratively until the stop condition (exceed max epoch or other conditions) is satisfied:

1) First, all the sample $\boldsymbol{x}_i^\Delta$ from training dataset will be transformed to feature vector $\boldsymbol{v}_i^\Delta$ by hidden layers.

$$\widetilde{\mathbb{D}^\Delta} = f(\mathbb{D}^\Delta) = \{(\boldsymbol{v}_1^\Delta, y_1^\Delta), (\boldsymbol{v}_2^\Delta, y_2^\Delta), \dots, (\boldsymbol{v}_n^\Delta, y_n^\Delta)\} \quad (11)$$

where $\boldsymbol{v}_i^\Delta = f(\boldsymbol{x}_i^\Delta)$, and $f(\cdot)$ is defined as equation (5).

2) The means (center) of the classes are computed in NCM layer by equation (3) using $\widetilde{\mathbb{D}^\Delta}$.

3) **for** mini_batch **in** $\widetilde{\mathbb{D}^\Delta}$:
   i. Compute the distances $d_{ik}$ by equation (6)
   ii. Compute the loss function $\mathcal{L}(\cdot)$ by equation (10).

iii. Leverage the stochastic gradient descent (SGD) optimization to fine-tune the parameters of hidden layers.

The implementation of initial training is summarized in **Algorithm 1**.

---

**Algorithm 1:** initial training algorithm of DNCM
**Input:**
The training dataset $\mathbb{D}^\Delta = \{(x_1^\Delta, y_1^\Delta), (x_2^\Delta, y_2^\Delta), \ldots, (x_n^\Delta, y_n^\Delta)\}$
where $x_i^\Delta \in \mathbb{R}^D$ is a signal of gas sensor of an odor,
$y_i^\Delta \in \{1, \cdots, K\}$ is the class label.
**output:**
parameters of hidden layers: **W**
**Initial:**
parameter of feature extractor: $\mathbf{W} = [W_0, \cdots, W_{h-1}]$
learning rate: $\delta \in (0, 0.1)$
momentum: $\gamma \in (0,1)$
velocity: $t = 0$
**Procedure:**
*while* epoch < max_epoch *do*
   epoch = epoch + 1
  1) Compute feature vectors of training samples
   *for* i=1 *to* n *do*
     $v_i^\Delta = f(x_i^\Delta)$       ($f(\cdot)$ defined as (5))
  2) Compute the class mean in feature space
   *for* k=1 to K *do*
$$c_k^\Delta = \frac{1}{N_k^\Delta} \sum_{v_i^\Delta \in class\ k} v_i^\Delta$$
  3) *for* mini_bach $\mathbb{B}$ *in* $\widetilde{\mathbb{D}^\Delta}$
    i. *for* $v_i^\Delta$ *in* $\mathbb{B}$
      *for* k=1 *to* K
$$d_{ik} = d(v_i, c_k)$$
    ii. Compute the loss function $\mathcal{L}(\mathbb{B})$
$$\mathcal{L}(\mathbb{B}) = -\sum_{i=1}^{n} \sum_{k=1}^{K} t_{ik} \log \frac{e^{-d(v_i, c_k)}}{\sum_m e^{-d(v_i, c_m)}}$$
    iii. Using stochastic gradient descent method for fine-tune the parameter **W**
$$t = \gamma \cdot t + \delta \cdot \frac{\partial \mathcal{L}(\mathbb{B})}{\partial \mathbf{W}}$$
$$\mathbf{W} = \mathbf{W} - t$$
*end while*

---

After finishing the initial training phase, the parameters of DNCM are fixed from this point onwards and shared.

*B. Updating training*

The updating training is designed to deal with new classes. The samples of new classes is coming in an open-ended fashion, $(x_1, y_1), (x_2, y_2), \ldots$. The DNCM leverage the hidden layers get each $x_i$'s feature vector $v_i$, then NCM layer will use $v_i$ to update $c_{y_i}$ (the mean of the class $y_i$). The procedure can repeat with nearly zero cost, which makes the DNCM can seamlessly integrate new classes. The algorithm is summarized in **Algorithm 2**.

---

**Algorithm 2**: Updating training algorithm of DNCM
**Input:**
The incremental dataset $\mathbb{D} = \{(x_1, y_1), (x_2, y_2), \ldots \ldots\}$.
where $x_i \in \mathbb{R}^D$ is a signal of gas sensor of an odor,
$y_i \in \{K+1, \cdots, \}$ is the class label.
**Output:**
The mean of new classes: $\{c_{K+1}, c_{K+2}, \cdots\}$
**Initial:**
Mean of classes: $c_k = 0$
Number of samples of classes: $N_k = 0$
**Procedure:**
  *for* each $x_i$ *in* $\mathbb{D}$ *do*
    1) compute the feature vector $v_i$ of $x_i$
       $v_i = f(x_i)$ ($f(\cdot)$ is defined as (5))
    2) updating the mean of class $y_i$
       $c_{y_i} = \frac{N_{y_i}}{N_{y_i}+1} c_{y_i} + \frac{1}{N_{y_i}+1} v_i$
       $N_{y_i} = N_{y_i} + 1$

---

**V Experiments**

In this section, we experimentally evaluate the proposed Deep Nearest Class Mean (DNCM) model, and compare it with the related incremental algorithms, including the baseline NCM [10], K-Nearest Neighbors (KNN) method [9], Incremental Activity Learning Framework (IALF) [29], MULticlass Transfer Incremental LEarning (MULTIpLE) [30], and the two current state-of-the-art incremental learning methods — NCMML [11] and NCMF [13]. The parameters of the compared approaches were optimized by cross-validation on initial dataset. In the following, we first describe the data sets and the parameter setups used in our experiments. Then, we compare the proposed model with other methods mainly in three aspects: i) accuracy; ii) efficiency; iii) Effect of size of training dataset.

*A. Experimental Dataset and Parameters Setup*

All experimental data developed in this paper were measured by an electronic nose system named pen3 in our laboratory. The pen3 E-nose system is based on 10 different semi-conductive metal-oxide gas (MOS) sensor array positioned into a very small chamber. It operates with filtered, ambient air as a carrier-gas at a flow rate of 100-200 mL/min, sample-chamber temperature of 15-25 ℃, humidity 40%-80% and sensor-array operating temperature of 200-500 ℃. During the measurement, gas sensors' response output was recorded every second. For each measurement, the steady sensor response value was extracted, and as a result, a 10-dimensional raw data vector was recorded.

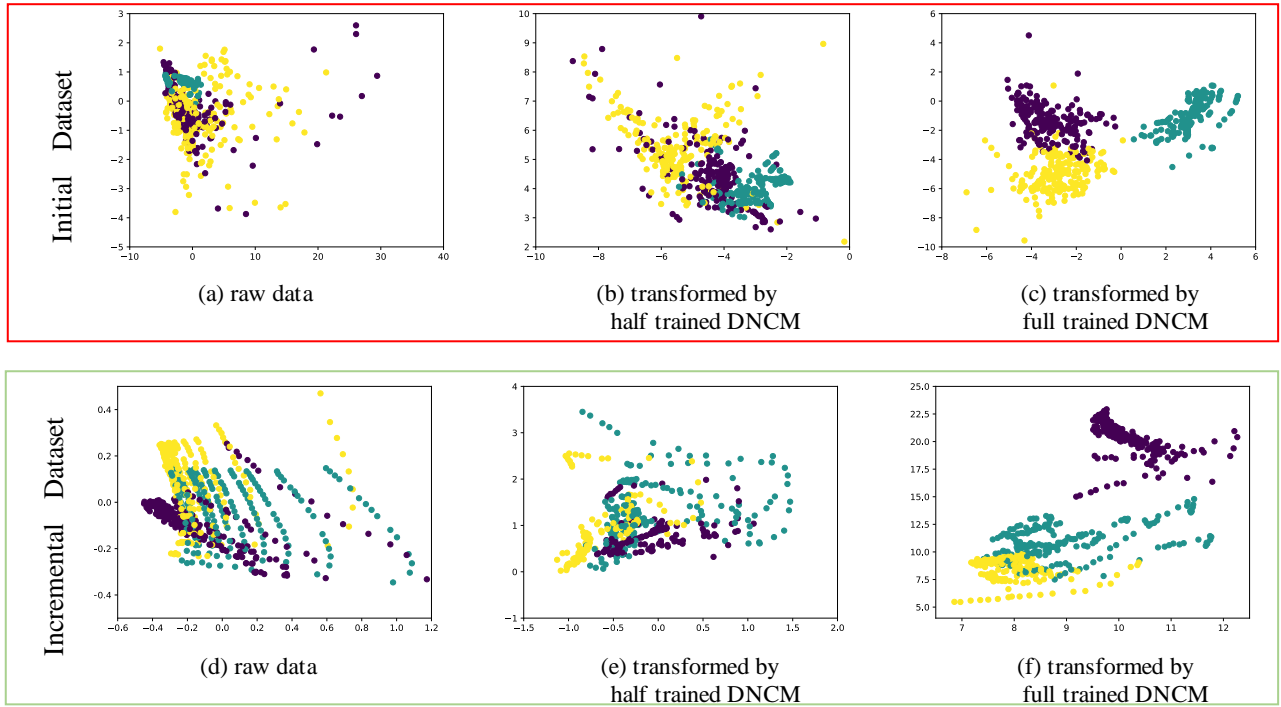

Fig. 3. We randomly choose data points of three classes from initial dataset and incremental dataset respectively. For visualization, the dimension of data points is reduced to 2 by PCA methods. (a), (b) and (c) show that data points in the initial dataset transformed by nothing, half and full trained neural network of DNCM respectively. So do (d), (e) and (f), except the data is from a different dataset – incremental dataset. The half (full) trained neural network (feature extractor) of DNCM means that the proposed model complete half (full) training iterations. From (a) to (c) show that the DNCM can find a feature space in which data points of different classes are separated. The accuracy of classification is much higher in this feature space than original data space. From (d) to (f) show that DNCM can produce the same effect on samples of new classes, although these samples have never been used to train the neural network of DNCM.

We divided the dataset into two parts: initial datasets and incremental dataset.

1) Initial Dataset: This dataset is a static and reasonable large dataset, which is used for initial training. It includes ten classes, and each class has 500 samples.

2) Incremental Dataset: Three types (Chinese herbs, fruits, textile), in which there are 25 specific classes odor, were collected in the incremented dataset. The detail of the initial dataset is shown in Table I. Like the initial dataset, each class also has 500 samples. In order to simulate real incremental odor recognition where new classes has a few amount of training samples, a small part of this dataset are used for training and the rest is used for testing.

In the experiments, we set the neural network in DNCM as 3 hidden layers, and the number of neurons in the three layers is 64, 32, and 20 respectively. The active function of each neuron is uniformly Rectified Linear Unit (ReLU). Although more advanced network structure may improve the results, the design and evaluation of network structure is not the focus of this paper. For NCM classifier of DNCM, we choose the Euclidean distance as the metric criterion.

## B. Training Protocol

The training process of DNCM is composed of two phases. In the initial training phase, the parameters of hidden layers are learned from the initial dataset. The incremental dataset is not available in this phase. 70%, 10% and 20% of data in the initial dataset were chosen for training, validation and testing, respectively. The initial training algorithm uses Stochastic Gradient Descent (SGD) for parameter optimization. The mini-batch size is 16, and momentum is set to 0.9. The initial learning rate is 0.001, which is decayed by 0.5 for 15 iterations. The max epoch is 50. In the updating training phase, the new class data from the incremental dataset will continue to valid in sequence with one sample at a time.

In the updating training phase, a small number of training samples were chosen from the incremental dataset, and the rest were used for testing. To avoid unstable experimental results caused by a few training samples, we set three experimental conditions. 1) The test set was relative **large**, with over 480 samples in each new class. 2) The training samples were **randomly** chosen from the incremental dataset at each experiment. 3) The experiment **repeated** 300 times, and the performance was measured using the average result. The experimental result was convincingly close to the real expectation by complying to these three conditions.

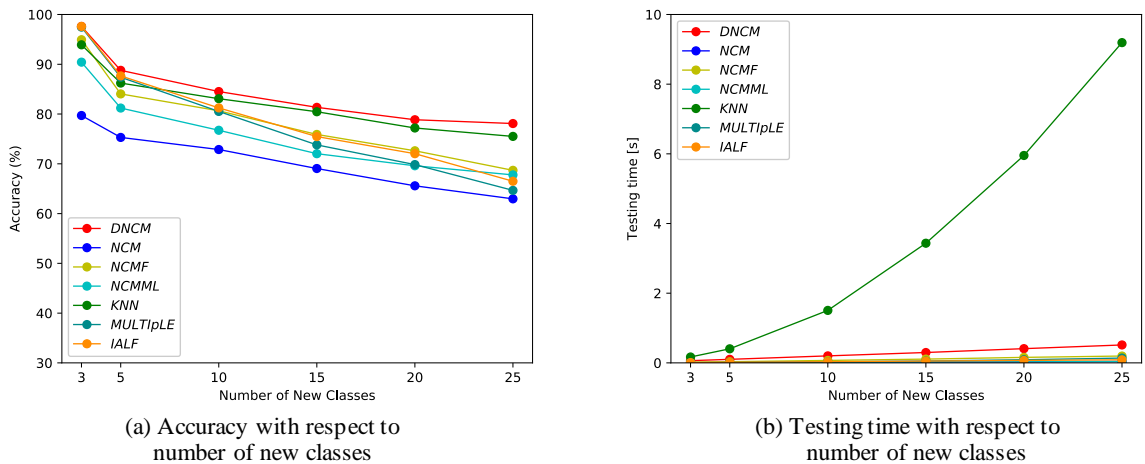

(a) Accuracy with respect to number of new classes

(b) Testing time with respect to number of new classes

Fig.4. Measurements at variable number of new classes for an incremental learning start with ten initial classes. (a) DNCM achieves 78.09 percent of accuracy at 25 new classes, which is better than the other compared methods. (b) The testing time of KNN increases rapidly with the number of classes, while that of DNCM (and other compared models) increases very slightly.

## C. Results

In this section, the experimental results are reported to evaluate the proposed DNCM method. We first observe the intuitive results in Fig. 3 which plots the scatter points of three different classes from initial and incremental dataset. Fig. 3 (a) and (d) show that raw data (signal of gas senor) of different classes are overlapped whether they are from initial or incremental dataset. Fig. 3 (b) and (e) show the feature vectors extracted by hidden layers of "half-trained" DNCM. The "half trained" DNCM means that it only completed half training iterations. Obviously, the overlap is reduced. Fig. 3 (c) and

(f) further show the feature vectors extracted by hidden layers of "fully trained" DNCM. The feature vectors have been almost completely separated. Fig. 3 (a)-(c) show that by training the proposed model, it can transform raw data of odor into a feature space where intra-class becomes more compact and the distance of inter-class becomes larger. Fig. 3 (d)-(f) show that the proposed model produces the same effect on new classes, although the data of new classes are not used to train the model.

In Fig. 4 we analyze the odor classification accuracy and the testing time with respect to different number of new classes from incremental dataset. The number of training samples in new classes is set to. 20. As seen in Fig. 4(a), when the number of new classes is less than 10, our proposed model DNCM has a little higher accuracy than other models. With the number of new classes increasing, the DNCM's advantage in term of accuracy becomes more apparent. When the number of new classes equals 25, the DNCM achieves 78.09% accuracy, which is 8-10% higher than NCMML [11] and NCMF [13], 11-13% higher than IALF [29] and MULTIpLE [30]. The reason is that in the feature space, by the aid of hidden layers, the inter-class distance is maximized, so that the distance between mean of new classes and mean of learned classes is relatively large. Then, integrating new class will not have a great impact on the region of learned classes.

We show the testing time in Fig. 4(b). The time complexity of DNCM, NCMF, NCMML, IALF, MULTIpLE and the baseline NCM are only related to the number of new classes, $K$, so it is O($K$). The testing time of KNN method is related to the number of samples ($N$). Its time complexity is O($N \times K$). Since $N \gg K$, its testing time grow rapidly as the number of class increase.

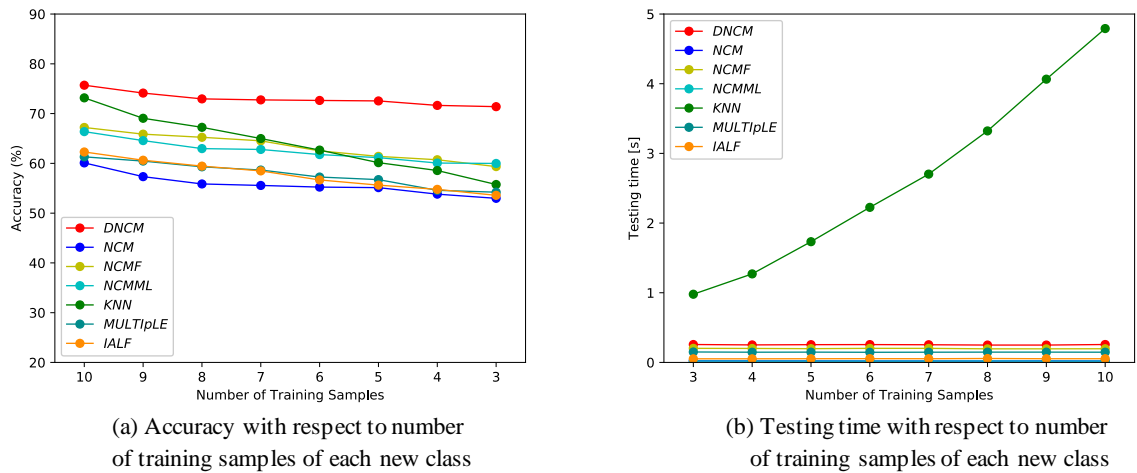

(a) Accuracy with respect to number of training samples of each new class

(b) Testing time with respect to number of training samples of each new class

Fig. 5. Measurements at a variable number of training samples of each new class. (a) shows that DNCM outperforms other methods in terms of accuracy on different numbers of training samples. Especially, when the number of samples equals 3, the recognition accuracy achieves 71.39%, which is 10% better than second place in comparisons. And DNCM is less sensitive to the decrease in the number of training samples than other compared methods. The accuracy of KNN dropped most significantly. (b) shows that testing time of KNN increases linearly with the number of samples.

The impact of the training data size ($N_k$) of new class $k$ is shown in Fig. 5, where fix the number of new classes $K = 25$. For $N_k$ from 10 to 3, DNCM outperforms other methods in term of accuracy, especially when $N_k$ is close to 3, as shown in Fig. 5 (a). The accuracy of DNCM archives 71.39%, while NCMML [11], NCMF [13], KNN, NCM, IALF [29] and MULTIpLE [30] achieves 59.98%, 59.36%, 55.75%, 52.98%, 53.58% and 54.20% respectively. Thus, DNCM improves more than 11% accuracy. In the feature space, by the aid of hidden layers, the intra-class compactness is minimized, so that reducing training samples of new class will not change the mean of new classes significantly. Moreover, because the NCM layer of DNCM uses the mean of classes to perform classification, the accuracy of the DNCM will not decline dramatically when the number of training samples of new classes decreases.

Fig. 5(b) show that the testing time of KNN increase linearly respect to the number of samples, and testing time of other models have little changed. This is because the testing of the models except KNN have nothing to do with the size of training set.

The detailed performance of the DNCM and compared methods on incremented dataset is shown in Table I. It can be seen that the DNCM provides better average accuracy.

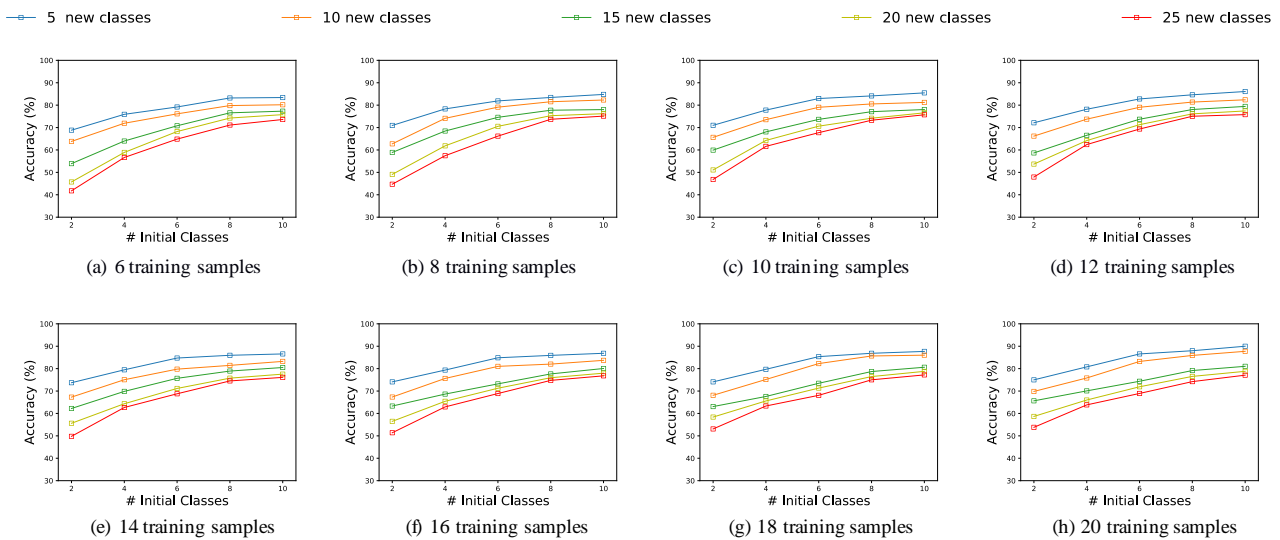

(a) 6 training samples  (b) 8 training samples  (c) 10 training samples  (d) 12 training samples
(e) 14 training samples  (f) 16 training samples  (g) 18 training samples  (h) 20 training samples

Fig. 6. Accuracy variation curves of different number of new classes with respect to the number of initial classes by frozen the number of training samples. (a) ~ (h) show the results of 6~20 training samples, respectively.

### D. Effect of size of training dataset

In the proposed DNCM model, there are two important settings of training dataset: the number of classes ($K_{init}$) in initial dataset and the number of classes ($K_{new}$) in incremental dataset. The impact of these two parameters on accuracy under different configurations is shown in Fig. 6. We can get two conclusions from the results. First, the accuracy of DNCM increases with $K_{init}$. This is consistent with the common logic of machine learning where the more training data results in more accurate model. Second, when the number of initial classes $K_{init}$ is small ($K_{init} \leq 4$), the number of new classes has greater impact on accuracy of DNCM. When $K_{init}$ is lager ($K \geq 8$), the number of new classes has relatively small impact. This means that increasing the number of classes in initial dataset can reduce the decline in accuracy due to the addition of new classes. This is because the ability of hidden layers to extract features is related to the number of class in initial dataset. The more classes in initial dataset, the ability of hidden layer to extract feature will be stronger. So as long as the number of class in initial dataset is enough, even if the number of new classes increase, the predicting accuracy of model will not decline significantly.

*E. Discussion*

In this paper, we have evaluated the algorithms from two aspects: testing time and classification accuracy. From the results shown in Fig. 4(a) and Fig. 5(a), it can be observed that the accuracy of DNCM is better than that of the two state-of-the-art incremental learning methods [11], [13], IALF [29], MULTIpLE [30] and other compared methods. KNN needs long testing time, while the DNCM, like other methods, requires relatively little testing time, which is illustrated in Fig. 4 (b) and Fig. 5 (b). The DNCM just needs to store the mean of these new classes and hidden layers. It requires only a small amount storage space. So it is suitable for a wide range of odor incremental classification applications, such as embedded odor recognition chip, real time odor classification equipment and so on.

TABLE I
Comparison of recognition accuracy on incremental dataset with 20 training samples and 10 initial classes

| Type | New Classes | DNCM | NCMML | NCMF | NCM | KNN | IALF | MULTIpLE |
|---|---|---|---|---|---|---|---|---|
| Chinese herbal | Musk | **99.85%** | 81.71% | 68.89% | 67.78% | 93.89% | 46.89% | 34.60% |
| | Centella | 66.11% | 46.29% | **75.56%** | 40.00% | 71.11% | 44.50% | 31.28% |
| | Microcos | 78.89% | 57.71% | 72.22% | 71.67% | 85.56% | 86.12% | **86.73%** |
| | sand ginger | **98.89%** | 84.00% | 75.56% | 81.67% | 87.22% | 50.24% | 41.23% |
| | Cardamom | 73.89% | 69.71% | 65.00% | 66.11% | 71.11% | 74.16% | **74.88%** |
| | Curcuma aromatica | 84.44% | 81.71% | 71.11% | 78.33% | 80.00% | 78.95% | **86.73%** |
| | Orthosiphon aristatus | 82.78% | **95.43%** | 92.78% | 91.11% | 93.33% | 83.78% | 83.84% |
| | Bombax ceiba | 66.11% | 84.00% | 87.22% | 83.89% | 86.67% | 79.00% | **87.68%** |
| | Eriobotrya japonica | **62.22%** | 30.29% | 64.44% | 28.33% | 53.89% | 39.57% | 38.53% |
| | Leonurus artemisia | **71.67%** | 40.00% | 48.89% | 28.89% | 53.89% | 36.36% | 35.07% |
| | Rabdosia serra | **88.33%** | 74.29% | 66.67% | 68.33% | 80.00% | 61.72% | 54.50% |
| | Atractylodes lancea | **94.44%** | 88.57% | 78.89% | 82.22% | 86.67% | 72.73% | 74.41% |
| | Paederia foetida | **95.56%** | 83.43% | 73.89% | 68.33% | 85.00% | 85.15% | 80.05% |
| Fruit | Dragon fruit | 70.00% | **75.43%** | 59.44% | 54.44% | 70.00% | 66.99% | 58.77% |
| | jujube | **58.89%** | 49.71% | 58.33% | 55.56% | 47.78% | 44.98% | 47.39% |
| | Mango | 79.44% | 29.71% | 59.44% | 25.00% | 71.67% | **84.71%** | 84.36% |
| | nectarine | **78.89%** | 61.14% | 61.67% | 54.44% | 67.22% | 57.89% | 52.13% |
| | Melon | 58.33% | 56.00% | 32.22% | 57.78% | **60.56%** | 59.33% | 57.82% |
| | orange | **94.44%** | 86.29% | 89.44% | 53.89% | 82.78% | 82.82% | 82.89% |
| | lemon | 76.67% | 54.86% | 79.44% | 56.11% | **88.33%** | 80.91% | 77.20% |
| | banana | 70.00% | **98.86%** | 56.11% | 86.11% | 86.11% | 87.61% | 87.63% |
| | tomato | 79.44% | 72.57% | 75.00% | 76.11% | **81.11%** | 80.91% | 81.00% |
| Textile materials | Mixed | **61.67%** | 46.86% | 47.78% | 53.89% | 47.22% | 44.35% | 43.74% |
| | Cotton | 81.11% | 74.29% | 68.89% | 72.78% | **89.44%** | 57.42% | 58.29% |
| | Wool | **80.00%** | 71.43% | 46.67% | 71.11% | 67.22% | 75.12% | 75.83% |
| Average | Average | *78.09%* | *67.77%* | *68.69%* | *62.96%* | *75.51%* | *66.49%* | *64.66%* |

# VI Conclusion

This paper proposes an incremental deep learning model, namely Deep Nearest Class Mean (DNCM) model for incremental multiclass odor classification. The proposed model presents a new framework based on the deep neural network model and the nearest class mean (DCM) method for handling the increasing classes over time.

The core idea of DNCM includes two aspects. One is to use deep neural network to find a feature space where different class samples are separated in this space. This feature space is valid not only for the initial dataset, but also for the data of new classes which are not used to train the model.

Second, to overcome the limitation that deep neural network (DNN) cannot efficiently handle the dynamically growing training dataset, we embed a NCM model as a layer into DNCM. The NCM layer can seamlessly integrate new classes. In addition, to handle dynamic dataset, we propose a two phase training algorithm that is different from DNN.

Finally, experiments were carried using the odor dataset and the performance of the proposed DNCM are compared with NCM [10], KNN [9], IALF [29] and MULTIpLE [30], two state-of-the-art incremental learning methods: NCMML [11] and NCMF [13]. The results show the DNCM is very efficient for incremental odor classification.